\pdfoutput=1
\documentclass[11pt]{article}

\usepackage[preprint]{acl}

\usepackage{times}
\usepackage{latexsym}

\usepackage[T1]{fontenc}

\usepackage[utf8]{inputenc}

\usepackage{microtype}

\usepackage{inconsolata}

\usepackage{graphicx}
\graphicspath{{../}{../pics/}{pics/}}

\usepackage{amsmath,amssymb}
\usepackage{enumitem}
\usepackage{algorithm}
\usepackage{algpseudocode}

\usepackage{booktabs}
\usepackage{multirow}
\usepackage{colortbl}
\usepackage{nicematrix}
\usepackage{tikz}
\usepackage{wrapfig}

\usepackage[most]{tcolorbox}

\definecolor{oursblue}{RGB}{222,235,247}
\definecolor{caseblue}{RGB}{235,244,252}
\definecolor{casegray}{RGB}{245,245,245}
\definecolor{casegreen}{RGB}{235,247,239}
\definecolor{caseorange}{RGB}{255,244,229}

\title{SkillLens: Adaptive Multi-Granularity Skill Reuse for Cost-Efficient LLM Agents}

\author{
  \textbf{Ziyang Yu}\textsuperscript{1,*}\quad
  \textbf{Yongliang Miao}\textsuperscript{1,*}\quad
  \textbf{Liang Zhao}\textsuperscript{1,2,\textdagger}\quad
  \textbf{Bowen Zhu}\textsuperscript{2}\quad
  \textbf{Hasibul Haque}\textsuperscript{2}
  \\
  \textsuperscript{1}Emory University\quad
  \textsuperscript{2}Casual Dynamics Lab
  \\
  \textsuperscript{*}Equal contribution\quad
  \textsuperscript{\textdagger}Corresponding author
}

\hypersetup{
  pdftitle={SkillLens: Adaptive Multi-Granularity Skill Reuse for Cost-Efficient LLM Agents},
  pdfauthor={Ziyang Yu, Yongliang Miao, Liang Zhao, Bowen Zhu, Hasibul Haque}
}

\begin{document}
\maketitle

\begin{abstract}
Skill libraries let LLM agents reuse procedural experience across tasks. However, existing systems treat skills as flat, single-resolution prompt blocks. This creates a tension between relevance and cost: injecting coarse skills can introduce irrelevant or misleading context, while rewriting entire skills is expensive and often unnecessary. We propose SkillLens, a hierarchical skill-evolution framework that organizes skills into a four-layer graph of policies, strategies, procedures, and primitives, and retrieves them at mixed granularity. Given a task, SkillLens first retrieves semantically relevant skill seeds, expands them through degree-corrected random walk over the skill graph, and then uses a verifier to decide whether each visited unit should be accepted, decomposed, rewritten, or skipped.
This lets the agent reuse compatible subskills directly while adapting only locally mismatched components. Over time, SkillLens further refines multi-granularity skills and the verifier's routing decisions. We provide theoretical analysis showing that mixed-granularity adaptation incurs sublinear cost under sparse mismatch assumptions and that the evolutionary update rule monotonically improves the validation objective until a local optimum. Across MuLocbench and ALFWorld, SkillLens consistently improves over strong skill-based baselines, achieving up to a 6.31 percentage-point Acc@1 gain for bug localization and raising agent success rate from 45.00\% to 51.31\%.
\end{abstract}
\vspace{-1.5em}
\section{Introduction}
\vspace{-0.5em}
Large language model (LLM) agents are increasingly used to solve tasks that require
multi-step interaction rather than single-shot prediction. In software engineering,
web navigation, tool use, data analysis, and interactive decision-making, agents must
plan over multiple steps, call external tools, inspect intermediate states, and revise
their behavior based on feedback~\citep{yao2023react,schick2023toolformer,wang2024codeact,madaan2023selfrefine}.
A recurring observation across these settings is that successful agent trajectories
often contain reusable procedural patterns: recurring ways of decomposing goals,
selecting tools, checking intermediate results, and recovering from failed actions.
Skill libraries have therefore emerged as a practical mechanism for externalizing such
procedural experience. Instead of relying on the model to rediscover useful behavior
from scratch for every new task, an agent can retrieve relevant prior skills and use
them as task-time guidance without retraining the underlying model.

Skill externalization falls into three broad lines: episodic
verbal-feedback memories such as Reflexion \citep{shinn2023reflexion}; flat skill
libraries that incrementally collect executable behaviors, exemplified by Voyager
\citep{wang2023voyager} and SkillAct \citep{liu2024skillact}; and natural-language
manuals or insights distilled from trajectories, such as ExpeL \citep{zhao2024expel} and
AutoManual \citep{chen2024automanual}. More recent systems push toward richer memory
infrastructures, including dynamically linked agentic memory \citep{xu2025amem},
procedural memory lifecycle management \citep{fang2025memp}, and self-evolving skill
ecosystems \citep{zhang2026evoskills}. Despite this progress, these systems share a
structural limitation: each skill or memory unit is treated as an atomic,
single-resolution entry. Retrieval is essentially flat lookup followed by full
verbatim insertion of the matched skill into the prompt. As a result, when a retrieved
skill is only partially compatible with the current task, the agent is forced into a
binary choice between accepting an imperfect block, which inflates context and seeds
hallucinations, or discarding it entirely and re-solving from scratch. This limitation also
appears in the bug-localization and repository-grounded agent literature
\citep{zhou2012buglocator,saha2013bluir,rahman2018blizzard,zhang2023repocoder,yang2024sweagent,zhang2024autocoderover,xia2024agentless,asad2025genloc},
where pipelines either commit to a single granularity (file-level ranking) or rely on
purely tool-driven exploration without a reusable procedural backbone, so cross-issue
regularities are repeatedly rediscovered at inference time.

This raises a precise question: at what resolution should a retrieved skill be reused?
A retrieved skill is rarely uniformly useful.
Its high-level plan may match the task while some concrete steps are irrelevant; conversely,
a low-level operation may transfer even when the surrounding strategy does not. Treating the
whole skill as the reuse unit therefore hides the actual source of transfer: the useful
part may live at a coarser level, a finer level, or inside a small component that only needs
local adaptation. We introduce SkillLens to make this resolution choice explicit.
Rather than assuming that skills are inherently trees, SkillLens uses a hierarchy as
an operational decomposition interface: it tells the agent where a retrieved skill can be
accepted as-is, expanded into finer units, locally rewritten, or skipped. Given a task,
SkillLens retrieves relevant skill seeds, expands them through the skill graph, and
uses a verifier to construct a compact task-specific skill context from units at mixed
granularities. The same view also changes skill evolution: failures should not only update
stored procedural content, but also refine the verifier's routing knowledge about when to
reuse, drill down, rewrite, or discard retrieved experience.

SkillLens couples inference-time adaptation with evolution-time refinement. Given a task, it retrieves semantically relevant
skill seeds, expands them through degree-corrected random walk over the skill graph, and
uses a verifier to route each visited unit to \textsc{Accept}, \textsc{Decompose},
\textsc{Rewrite}, or \textsc{Skip}. The final context is composed only from accepted and
locally rewritten units, so the agent receives a compact task-specific skill context rather
than a full retrieved skill block. SkillLens also extends this mechanism to
evolution: gap reports from failed trajectories refine both the agent-side procedural
registry and the verifier-side routing knowledge, improving not only what skills are stored
but also how they are reused. We evaluate SkillLens on MuLocbench and ALFWorld to
study mixed-granularity skill reuse in repository-level localization and long-horizon
interactive agent tasks.

In summary, our contributions are threefold:
\begin{enumerate}[leftmargin=10pt, topsep=-2pt, itemsep=1pt, partopsep=1pt, parsep=1pt]
    \item We formulate skill reuse as a context selection and adaptation problem: instead of directly injecting retrieved skills as flat prompt blocks, the agent should identify which parts of the retrieved  knowledge are useful for the current task and compose them into a compact skill context.

    \item We introduce SkillLens, a resolution-aware framework that builds skill contexts through hierarchical retrieval, verifier-guided decomposition, and localized adaptation. This allows partially relevant skills to be reused at the appropriate granularity rather than accepted or discarded wholesale.

    \item We provide theoretical and empirical evidence that SkillLens improves both efficiency and performance. We show sublinear adaptation cost under sparse mismatch assumptions and monotone improvement of the evolution objective; empirically, SkillLens improves MuLocbench Acc@1 by up to 6.31 points and raises ALFWorld success rate from 45.00\% to 51.31\%.
\end{enumerate}

\vspace{-1mm}
\section{Related Work}
\label{sec:related-work}
\vspace{-2mm}
\paragraph{Skill libraries as procedural memory.}
LLM agents increasingly externalize experience as reusable procedural memory. Early agent frameworks such as ReAct~\citep{yao2023react}, Toolformer~\citep{schick2023toolformer}, CodeAct~\citep{wang2024codeact}, and Self-Refine~\citep{madaan2023selfrefine} combine reasoning, tool use, and feedback to solve interactive tasks. Building on this, Reflexion~\citep{shinn2023reflexion}, Voyager~\citep{wang2023voyager}, SkillAct~\citep{liu2024skillact}, ExpeL~\citep{zhao2024expel}, AutoManual~\citep{chen2024automanual}, and Skill-Pro~\citep{mi2026skillpro} store reflections, executable skills, distilled insights, or reusable procedures. Recent surveys further formalize agent skills as modular packages of instructions, code, and resources that can extend agent capabilities without retraining~\citep{xu2026agentskills}. However, most skill libraries still treat a skill as a flat unit that is retrieved and inserted as a whole, which makes partial reuse difficult; SkillLens instead represents skills as hierarchical procedural objects and constructs contexts from selected sub-units.
\vspace{-2mm}
\paragraph{Evolving agent memory and skill systems.}
Another line studies how agent memories and skills evolve. MemoryBank~\citep{zhong2023memorybank}, A-MEM~\citep{xu2025amem}, Mem0~\citep{chhikara2025mem0}, and Memp~\citep{fang2025memp} organize and revise long-term or procedural memories. Skill-evolution systems such as AutoSkill~\citep{yang2026autoskill}, EvoSkill~\citep{alzubi2026evoskill}, and EvoSkills~\citep{zhang2026evoskills} use interaction traces, failure analysis, or co-evolutionary verification to refine reusable skill packages while keeping the base model fixed. These methods show that external skills can evolve through feedback, but they mainly optimize the stored skill content itself, while the decision of how to reuse a retrieved skill remains implicit or coarse-grained. SkillLens addresses this limitation with a dual-registry design that evolves both the agent-side procedural registry and the verifier-side routing knowledge.
\vspace{-2mm}
\paragraph{Skill routing and context construction.}
As skill libraries scale, selecting useful context becomes a key bottleneck. 
SkillRouter~\citep{zheng2026skillrouter} shows that large-scale skill routing requires more than shallow metadata matching, while recent benchmarks find that skill use becomes fragile when agents retrieve from large, noisy collections or receive imperfectly matched skills~\citep{liu2026agenticskillswild}. 
More broadly, memory-augmented agents rely on retrieved context for long-horizon reasoning and decision-making~\citep{zhang2024memorysurvey,du2026memoryagents}. 
However, most approaches focus on which skill or memory to retrieve, while still consuming the selected context at a single resolution. 
SkillLens addresses this limitation by constructing mixed-granularity skill contexts after retrieval, enabling useful subunits to be accepted, decomposed, or adapted under token and computation budgets.

\vspace{-2mm}

\section{Methodology}

\subsection{Problem Formulation}
\label{sec:problem-formulation}
\paragraph{Setup.}
We study skill-mediated problem solving for a language agent \(\mathcal{A}_{\theta}\) 
acting on tasks \((Q,Y)\sim\mathcal{D}\), where \(Q\) is a problem instance and \(Y\) its
ground-truth answer or successful trajectory. The agent has access to a skill library
\(\mathcal{S}\) of textual procedural units. Given \(Q\), a retrieval
and adaptation procedure constructs a skill context \(\mathcal{K}(Q;\mathcal{S})\) from
\(\mathcal{S}\) and augments the agent's prompt with \(\mathcal{K}\); the agent then produces
an output \(O_{\mathcal{S}}(Q)=\mathcal{A}_{\theta}\!\left(Q,\mathcal{K}(Q;\mathcal{S})\right)\).
Task quality is measured by a metric \(M\!\left(Y,O_{\mathcal{S}}(Q)\right)\in[0,1]\), and the
cost of constructing and consuming \(\mathcal{K}\) (token count for retrieval, recursive verification,
localized rewriting, and final inference) is summarized by a normalized cost \(\mathrm{C}(Q,\mathcal{S})\).

{\setlength{\parskip}{0pt}%
\vspace{4pt}%
\noindent\textbf{Hierarchical skill graph.}
We organize \(\mathcal{S}\) into a four-layer skill graph \(\mathcal{G}=(\mathcal{S},\mathcal{E})\) 
that adapts temporal-abstraction hierarchies from the options framework~\citep{sutton1999options}, 
feudal manager--worker designs~\citep{vezhnevets2017feudal}, three-layer robotic architectures~\citep{gat1998threelayer}, 
and HTN planning~\citep{erol1994htn} to skill-mediated LLM agents~\citep{yao2023react,wang2023voyager}. 
Each skill \(s\in\mathcal{S}\) carries a tuple \((\ell(s),\mathrm{ch}(s),\tau(s))\) with layer index \(\ell(s)\!\in\!\{1,2,3,4\}\), 
an ordered set of children \(\mathrm{ch}(s)\) realizing \(s\) at the next finer layer, and textual procedural content \(\tau(s)\); 
the four values of \(\ell\) name the abstraction layers: 
(1) \(\ell{=}1\) (\emph{Policy}): routes a task to a downstream strategy;
(2) \(\ell{=}2\) (\emph{Strategy}): gives long-horizon plans answering \emph{how to solve overall};
(3) \(\ell{=}3\) (\emph{Procedure}): provides reusable SOPs as ordered step sequences with explicit branching and retry;
(4) \(\ell{=}4\) (\emph{Primitive}): holds atomic executable units (tool calls, environment actions) that are not further decomposed~\citep{schick2023toolformer}.

\looseness=-1
Hierarchical edges \(\mathcal{E}_{\mathrm{h}}\) connect only adjacent layers (\(\ell(s_j)=\ell(s_i)+1\)) 
along \(\mathrm{ch}(\cdot)\), while lateral edges \(\mathcal{E}_{\mathrm{lat}}\) connect siblings 
within a layer; weights combine semantic and structural signal, 
\(e_{ij}=\mathrm{sim}_{\mathrm{emb}}(s_i,s_j)\cdot w_{ij}\), with \(\mathrm{sim}_{\mathrm{emb}}\) 
the embedding cosine similarity and \(w_{ij}\) determined by whether \(s_i,s_j\) are in a parent--child 
or sibling relation.

\vspace{4pt}%
\noindent\textbf{Optimization objective.}
We frame the design of the skill system as a optimization problem over the skill registry.
Let \(\mathcal{D}_{\mathrm{ev}}\) be the \emph{evolution split}, the data used to drive iterative updates of the skill registry. It is analogous to a training split, but consumed for registry evolution rather than gradient-based weight updates, since \(\mathcal{A}_{\theta}\) is a frozen LLM. We seek
\setlength{\belowdisplayskip}{4pt}%
\setlength{\abovedisplayskip}{4pt}%
\setlength{\abovedisplayshortskip}{4pt}%
\setlength{\belowdisplayshortskip}{4pt}%
\begin{align}
  \mathcal{S}^{*}
&=
\arg\max\nolimits_{\mathcal{S}}\;
J(\mathcal{S}),
\label{eq:objective}\\
J(\mathcal{S})
&:=
\mathop{\mathbb{E}}\limits_{(Q,Y)\sim\mathcal{D}_{\mathrm{ev}}}
\!\left[
M\!\big(Y, O_{\mathcal{S}}(Q)\big)
- \mathrm{C}(Q, \mathcal{S})
\right].
\nonumber
\end{align}}
\looseness=-1
The metric \(M\!\big(Y,O_{\mathcal{S}}(Q)\big)\) rewards output correctness, while
\(\mathrm{C}(Q,\mathcal{S})\) penalizes the inference-time resources used to produce
\(O_{\mathcal{S}}(Q)\). Section~\ref{sec:skill-graph-adaptation} addresses zero-shot
adaptation, where \(M\) is unavailable at query time, by retrieving and routing over
multi-granularity skills from a fixed \(\mathcal{S}\). Section~\ref{sec:skill-verifier-evolution}
uses \(M\) on the evolution split to jointly refine the skills and \(\mathcal{S}\) under
a single iterative update rule.


\vspace{0pt}

\subsection{Hierarchical Skill Graph and Mixed-Granularity Adaptation}
\label{sec:skill-graph-adaptation}
\vspace{0pt}
\looseness=-1
Our key insight is that skill compatibility is rarely uniform across a retrieved skill:
parts of a skill are reusable as-is, parts need to be expanded to a finer resolution,
parts need to be locally rewritten, and parts should be discarded. We exploit the four-layer structure
of \(\mathcal{G}\) to make each of these decisions \emph{locally}, on the smallest necessary unit, and
we surface procedurally related skills via a degree-corrected random walk that resists collapse into
dense similarity clusters. Concretely, a retrieved high-level skill is not an indivisible prompt
block but a subtree of \(\mathcal{G}\): a policy node expands into strategy children, a strategy expands
into procedures, and a procedure expands into primitives. This recursive structural mapping turns each
retrieved skill into a procedural object addressable at multiple resolutions, and the remainder of this
section operates on this subtree.

\noindent \textbf{Per-query surrogate objective.}
At fixed registries, Eq.~\eqref{eq:objective} decouples across queries through \(\mathcal{K}(Q)\), and the routing \(\{a_u\}\) of Eq.~\eqref{eq:routing} fully determines \(\mathcal{K}(Q)\) via Eq.~\eqref{eq:context}. Let \(R=\{u:a_u=\textsc{Accept}\}\cup\{u:a_u=\textsc{Rewrite}\}\subseteq\mathcal{S}\) be the retained units and \(B=\{u:a_u=\textsc{Rewrite}\}\subseteq R\) the rewritten subset. Since \(M\) is unobservable at query time and the retrieval term is combinatorial, we replace \(M\) by the tractable surrogate
{\setlength{\abovedisplayskip}{4pt}%
\setlength{\belowdisplayskip}{4pt}
\begin{equation}
\label{eq:surrogate}
\widehat{U}_Q(R,B)\;=\;F_Q(R)\;+\;\sum\nolimits_{u\in B}\Delta^{\mathrm{rw}}_Q(u),
\end{equation}
}
where \(F_Q\) is a monotone submodular query--skill relevance aggregator and \(\Delta^{\mathrm{rw}}_Q(u)\) is the predicted utility gain from rewriting \(u\) (additive because \(\mathcal{W}_{\phi}\) acts on each unit independently). The per-query objective \(\widehat{U}_Q(R,B)-\mathrm{C}(Q,\mathcal{S})\) thus separates into a retrieval block (in \(R\)) and a rewrite block (in \(B\), as a tree dynamic program over the decomposition subtree). Algorithm~\ref{alg:adaptation} approximates the two blocks in lines~\ref{alg:line:trigger-start}--\ref{alg:line:partition} and lines~\ref{alg:line:traverse-start}--\ref{alg:line:traverse-end} respectively; the composed guarantee is Proposition~\ref{prop:composed} (proof in Appendix~\ref{app:proof-prop3}).

{%
\setlength{\intextsep}{2pt}%
\setlength{\floatsep}{2pt}%
\setlength{\textfloatsep}{2pt}
\begin{algorithm}[tbp]
\small
\caption{Mixed-Granularity Skill Adaptation}
\label{alg:adaptation}
\begin{algorithmic}[1]
\Require Query $Q$; skill graph $\mathcal{G}$; agent $\mathcal{A}_{\theta}$; verifier $\mathcal{V}_{\phi}$; writer $\mathcal{W}_{\phi}$
\Ensure Skill context $\mathcal{K}(Q)$
\Statex \textbf{Stage 1: Trigger, seed retrieval, and RWR (standard components).}
\State \textbf{if} confidence of $\mathcal{A}_{\theta}(Q)$ exceeds threshold \textbf{then return} $\varnothing$ \label{alg:line:trigger-start}
\State Retrieve top-$K$ embedding seeds $\mathbf{p}_0$ over $\mathcal{S}$
\State Run degree-corrected RWR on $\mathcal{G}$ from $\mathbf{p}_0$ to obtain candidate scores $\mathbf{s}$
\State Partition candidates by $\mathbf{s}$ into $\mathcal{C}_{\mathrm{full}},\mathcal{C}_{\mathrm{part}},\mathcal{C}_{\mathrm{mis}}$; set $\mathcal{K}\gets\mathcal{C}_{\mathrm{full}}$ \label{alg:line:partition}
\Statex \textbf{Stage 2: Verifier-driven recursive traversal (proposed).}
\For{each root $r \in \mathcal{C}_{\mathrm{part}}$} \label{alg:line:traverse-start}
  \State Initialize stack $\mathcal{Q} \gets \{r\}$
  \While{$\mathcal{Q} \ne \varnothing$}
    \State $u \gets \mathrm{pop}(\mathcal{Q})$;\quad $a_u \gets \mathcal{V}_{\phi}(Q,u)$
    \If{$a_u = \textsc{Accept}$} $\mathcal{K} \gets \mathcal{K} \cup \{u\}$
    \ElsIf{$a_u = \textsc{Decompose}$} push children $\mathrm{ch}(u)$ onto $\mathcal{Q}$
    \ElsIf{$a_u = \textsc{Rewrite}$} $\mathcal{K} \gets \mathcal{K} \cup \{\mathcal{W}_{\phi}(Q,u)\}$
    \ElsIf{$a_u = \textsc{Skip}$} \textbf{continue}
    \EndIf
  \EndWhile
\EndFor \label{alg:line:traverse-end}
\State \Return $\mathrm{Compose}(\mathcal{K})$
\end{algorithmic}
\end{algorithm}
}

\noindent\textbf{Compatibility partition and verifier-driven traversal.}
The scores \(\mathbf{s}\) partition retrieved candidates into three tiers via calibrated intervals:
\emph{fully compatible} skills, which are injected directly; \emph{partially relevant} skills, which
require recursive adaptation; and \emph{mismatched} skills, which are discarded.
For partially relevant skills, full-scale rewriting conflates compatible substructure with local mismatches
and inflates token cost. Instead, we traverse the skill's decomposition subtree top-down, 
expanding a node only when its utility cannot be decided at the current granularity. The verifier emits, for each visited unit \(u\), a compact routing action:
{\setlength{\abovedisplayskip}{4pt}%
\setlength{\belowdisplayskip}{4pt}
\begin{equation}
\label{eq:routing}
\begin{aligned}
a_u = \mathcal{V}_{\phi}(Q,u)\in\{&\textsc{Accept},\,\textsc{Decompose},\\
&\textsc{Rewrite},\,\textsc{Skip}\}.
\end{aligned}
\end{equation}}
\textsc{Accept} retains an informative compatible unit and terminates traversal of its subtree; 
\textsc{Decompose} expands the unit into its lower-level children for further evaluation; \textsc{Rewrite} 
locally adapts a useful but mismatched unit via the writer \(\mathcal{W}_{\phi}\); and \textsc{Skip} excludes 
the unit. The set of nodes terminated by \textsc{Accept} or \textsc{Skip} forms an \emph{adaptive frontier} 
over the hierarchy. The frontier is consolidated into the final context:
{\setlength{\abovedisplayskip}{4pt}%
\setlength{\belowdisplayskip}{4pt}
\begin{equation}
\label{eq:context}
\begin{aligned}
\mathcal{K}(Q)
&=
\mathrm{Compose}\big(\{u:a_u=\textsc{Accept}\}\\
&\quad\cup\{\mathcal{W}_{\phi}(Q,u):a_u=\textsc{Rewrite}\}
\big),
\end{aligned}
\end{equation}}
which is then concatenated with the prompt for a second reasoning pass by \(\mathcal{A}_{\theta}\), 
yielding the final output. Token and computation cost scale with the number of visited and rewritten units,
not the size of the retrieved skill tree. Algorithm~\ref{alg:adaptation} summarizes the procedure: 
the first stage (lines~\ref{alg:line:trigger-start}--\ref{alg:line:partition}) reuses standard components, namely confidence-gated triggering, embedding-based top-$K$ retrieval, and degree-corrected random walk with restart, to produce a scored candidate set partitioned into compatibility tiers, and the second stage 
(lines~\ref{alg:line:traverse-start}--\ref{alg:line:traverse-end}) is the proposed verifier-driven recursive 
traversal over the decomposition subtree of each partially relevant candidate.
\subsection{Dual-Registry Evolutionary Refinement}
\label{sec:skill-verifier-evolution}
\vspace{-1mm}
We separate the skill registry of Section~\ref{sec:problem-formulation} into 
two coupled components: the \emph{agent registry} \(\mathcal{S}\), 
the procedural knowledge consulted by \(\mathcal{A}_{\theta}\), and the \emph{verifier registry} 
\(\mathcal{S}_{\mathcal{V}}\), which parameterizes the routing policy of 
(when to \textsc{Accept}/\textsc{Decompose}/\textsc{Rewrite}/\textsc{Skip}). The inference 
procedure of Section~\ref{sec:skill-graph-adaptation} treats both as fixed: localized rewrites by 
\(\mathcal{W}_{\phi}\) act on retrieved skill instances for the current query and are not written back 
into \(\mathcal{S}\). We therefore approach the joint objective in Eq.~\eqref{eq:objective} 
via an iterative dual-registry update rule that first updates \(\mathcal{S}_{\mathcal{V}}\) and
then derives \(\mathcal{S}\) from it.

\noindent \textbf{Gap-report construction.}
For each task in the evolution split, we run the mixed-granularity adaptation 
protocol of Section~\ref{sec:skill-graph-adaptation} to obtain 
\(O_{\mathcal{S}}(Q)\), 
then compare it against the ground truth \(Y\) to construct a \emph{gap report}. 
The report records (i) the error type, (ii) the retrieved skills and recursive 
traversal path, (iii) the verifier's routing actions \(\{a_u\}\), and (iv) the 
units invoked or rewritten. This isolates whether a failure stems from missing/incorrect 
agent-side procedural knowledge or from miscalibrated verifier-side routing, and supplies 
the writer \(\mathcal{W}_{\phi}\) with error-conditioned signals for proposing edits.
{\setlength{\intextsep}{2pt}%
\setlength{\floatsep}{2pt}%
\setlength{\textfloatsep}{2pt}%
\begin{algorithm}[h]
\caption{Dual-Registry Evolutionary Refinement (one iteration)}
\label{alg:evolution}
\begin{algorithmic}[1]
\Require Registries $(\mathcal{S}^{(t)},\mathcal{S}_{\mathcal{V}}^{(t)})$; split $\mathcal{D}_{\mathrm{ev}}$; $\mathcal{A}_{\theta},\mathcal{V}_{\phi},\mathcal{W}_{\phi}$; objective $J$
\Ensure Updated $(\mathcal{S}^{(t+1)},\mathcal{S}_{\mathcal{V}}^{(t+1)})$
\State $\mathcal{R}_t \gets \varnothing$ \label{alg:evo:gap-start}
\For{$(Q,Y)\in\mathcal{D}_{\mathrm{ev}}$}
  \State Run Alg.~\ref{alg:adaptation} to get $O_{\mathcal{S}}(Q)$ and trace $\mathcal{T}_Q$; if $M(Y,O_{\mathcal{S}}(Q))<1$, add $(\text{err},\mathcal{T}_Q)$ to $\mathcal{R}_t$
\EndFor \label{alg:evo:gap-end}
\State $\mathcal{P}_t \gets \{(\mathcal{S}^{(t)},\mathcal{S}_{\mathcal{V}}^{(t)})\}$ \Comment{fallback} \label{alg:evo:propose-start}
\For{$\mathfrak{r}\in\mathcal{R}_t$, target $\mathfrak{t}$ inferred from $\mathfrak{r}$, $\mathrm{op}\in\{\textsc{Add},\textsc{Del},\textsc{Upd},\textsc{Mrg}\}$}
  \State $\mathcal{P}_t \mathrel{+}= \{\mathcal{W}_{\phi}(\mathrm{op},\mathfrak{t},\mathfrak{r};\mathcal{S}^{(t)},\mathcal{S}_{\mathcal{V}}^{(t)})\}$
\EndFor \label{alg:evo:propose-end}
\State \Return $\arg\max_{(\mathcal{S},\mathcal{S}_{\mathcal{V}})\in\mathcal{P}_t} J(\mathcal{S},\mathcal{S}_{\mathcal{V}})$ \label{alg:evo:commit}
\end{algorithmic}
\end{algorithm}}

\noindent \textbf{Coupled update rule.}
The two registries are not optimized independently: \(\mathcal{S}_{\mathcal{V}}\) is updated
first, and the agent registry \(\mathcal{S}\) is then determined by it. Given a verifier
registry \(\mathcal{S}_{\mathcal{V}}\), the resulting routing decisions
(\textsc{Accept}/\textsc{Decompose}/\textsc{Rewrite}/\textsc{Skip}) over trajectories on
\(\mathcal{D}_{\mathrm{ev}}\) specify which units are kept, decomposed, or rewritten, and
\(\mathcal{W}_{\phi}\) materializes the corresponding agent-side edits. We therefore treat the next agent registry as an induced output
\(\Phi(\mathcal{S}^{(t)},\mathcal{S}_{\mathcal{V}})\), where \(\Phi\) denotes the agent-side
edits applied by \(\mathcal{W}_{\phi}\) to the current \(\mathcal{S}^{(t)}\) under the
trajectories that \(\mathcal{S}_{\mathcal{V}}\) routes on \(\mathcal{D}_{\mathrm{ev}}\). At iteration \(t\), gap reports drive
\(\mathcal{W}_{\phi}\) to propose a finite set of single-operator edits to
\(\mathcal{S}_{\mathcal{V}}\), which in turn induces a finite candidate set
\(\mathcal{C}_t=\{\Phi(\mathcal{S}^{(t)},\mathcal{S}_{\mathcal{V}}):\mathcal{S}_{\mathcal{V}}\in\mathcal{P}_t\}\)
of agent registries (with \(\mathcal{S}^{(t)}\in\mathcal{C}_t\) as a fallback). We then commit
the candidate that maximizes the evolution objective \(J\) of
Eq.~\eqref{eq:objective}:
{\setlength{\abovedisplayskip}{4pt}%
\setlength{\belowdisplayskip}{4pt}
\begin{equation}
\label{eq:update}
\mathcal{S}^{(t+1)}
\;=\;
\arg\max\nolimits_{\mathcal{S}\,\in\,\mathcal{C}_t}
J(\mathcal{S}^{(t)}, \mathcal{S}_{\mathcal{V}}^{(t)}).
\end{equation}}
Verifier-side edits sharpen routing so that the synthesized context fed to \(\mathcal{A}_{\theta}\)
is more useful, and the induced agent-side edits via \(\Phi\) directly improve the procedural
knowledge \(\mathcal{A}_{\theta}\) consults. Because both updates are driven by the same writer
from a unified gap report, the two registries co-evolve on the same error signal. In summary,
one evolution iteration proceeds in three steps: gap reports are built from failed
evolution-split trajectories; the writer proposes a finite candidate set of single-operator
edits to either registry; and the pair maximizing the evolution objective $J$ is committed,
with the current pair retained as a fallback.
\subsection{Theoretical Analysis}
\label{sec:theoretical-analysis}
\vspace{-1mm}
We provide two complementary results that justify our design: (i) the inference-time mixed-granularity
adaptation incurs sublinear cost in the size of the retrieved skill tree, and (ii) the evolution-stage
updates yield monotone improvement and converge to a local optimum of the evolution objective.
The proof is deferred to Appendix~\ref{app:proofs}.

\noindent\textbf{Proposition 1 (Sublinear cost of adaptation).}
Let the retrieved skill tree \(T\) (the four-layer hierarchy of
Section~\ref{sec:skill-graph-adaptation}) have \(n\) nodes, maximum branching factor \(b\), and
depth \(D=O(\log_b n)\), and suppose the verifier routes only a small fraction \(\rho<1/b\) of
subtrees to \textsc{Decompose} (the remainder being \textsc{Accept}ed, locally
\textsc{Rewrite}n, or \textsc{Skip}ped). Let \(N_{\mathrm{vis}}\) be the number of nodes
visited by the verifier-driven traversal and \(N_{\mathrm{rw}}\) the number of
\textsc{Rewrite} actions on \(T\). Then the per-task adaptation cost satisfies
\(\mathrm{C}_{\mathrm{adapt}}=o(n)\).


\noindent\textbf{Proposition 2 (Monotone improvement and convergence to a local optimum).}
Let \(J\) be the evolution objective in Eq.~\eqref{eq:objective} and let \(\mathcal{S}^{(t)}\)
evolve under the update rule in Eq.~\eqref{eq:update}. Since \(M\in[0,1]\) and
\(\mathrm{C}\ge 0\), \(J\) is bounded above; the candidate set \(\mathcal{C}_t\) always
includes \(\mathcal{S}^{(t)}\) as a ``do nothing'' fallback (Eq.~\eqref{eq:update}). Then the
sequence \(\{J^{(t)}\}_{t\ge 0}\) with \(J^{(t)}=J(\mathcal{S}^{(t)})\) is non-decreasing and
converges to a value \(J^{\dagger}\le J^{*}\), where
\(J^{*}=\max_{\mathcal{S}}J(\mathcal{S})\). Moreover, the limit point
\(\mathcal{S}^{\dagger}\) is a local optimum in the sense that no atomic operator applied to
either registry strictly increases \(J\).


\noindent\textbf{Proposition 3 (Composed approximation guarantee).}\label{prop:composed}
Suppose: (a) the retrieval scores \(r_Q(u)\) calibrate the per-skill utility \(F_Q(\{u\})\) up to error \(\epsilon_{\mathrm{ret}}\); (b) \(F_Q\) is monotone submodular; (c) the degree-corrected RWR score \(s(u)\) estimates the marginal gain \(\Delta(u\mid R)\) up to error \(\epsilon_{\mathrm{rwr}}\); and (d) the verifier matches the optimal local action of Eq.~\eqref{eq:routing} up to error \(\epsilon_{\mathrm{ver}}\). Let \((\widehat{R},\widehat{B})\) be the retrieved-and-rewritten pair returned by Algorithm~\ref{alg:adaptation} and \((R^{*},B^{*})\) the optimum of the surrogate \(J_Q\) (the objective of Eq.~\eqref{eq:objective} with the unobservable correctness term replaced by its tractable surrogate). Then
\begin{equation}
\label{eq:composed-bound}
\begin{aligned}
J_Q(\widehat{R},\widehat{B})
&\ge
(1-1/e)\,J_Q^{\mathrm{ret}}(R^{*})+J_Q^{\mathrm{rw}}(B^{*})\\
&\quad-\bigl(\epsilon_{\mathrm{ret}}+\epsilon_{\mathrm{rwr}}+|\widehat{R}|\,\epsilon_{\mathrm{ver}}\bigr).
\end{aligned}
\end{equation}


In practice, two factors tighten the sub-optimality gap: (i) the gap report supplies
error-conditioned proposals, biasing \(\mathcal{P}_t\) toward edits that are likely to yield 
large positive \(\Delta J\); and (ii) the four operators are jointly closed under composition 
over the registry space (any registry pair is reachable via a finite sequence of 
\textsc{Add}/\textsc{Delete}/\textsc{Update}/\textsc{Merge} operations), so the only obstruction 
to global optimality is the greedy single-step commit rule rather than the expressiveness of the 
action set.

\section{Experiment}


\begin{table*}[t]
\centering
\small
\caption{MuLocbench results across different granularity levels. All values are reported in percentage. Better values are in bold.}
\vspace{4pt}
\label{tab:combined_results}
\setlength{\tabcolsep}{4.2pt}
\renewcommand{\arraystretch}{1.08}
\resizebox{\textwidth}{!}{
\begin{NiceTabular}{cccccc|cccc|cccc}

\toprule
\multirow{2}{*}{\textbf{Granularity}}
& \multirow{2}{*}{\textbf{Method}}
& \multicolumn{4}{c|}{\textbf{Accuracy}}
& \multicolumn{4}{c|}{\textbf{Recall}}
& \multicolumn{4}{c}{\textbf{Precision}} \\
\cmidrule(r){3-6} \cmidrule(lr){7-10} \cmidrule(l){11-14}
& & @1 & @3 & @5 & @10
& @1 & @3 & @5 & @10
& @1 & @3 & @5 & @10 \\
\midrule

\multirow{4}{*}{\textbf{File}}
& Baseline
& 58.87 & 70.57 & 72.45 & 72.83
& 31.82 & 46.39 & 49.08 & 49.62
& 58.87 & 49.06 & 48.21 & 47.97 \\
& AutoSkill
& 58.58 & 71.70 & 72.78 & 73.37
& 34.33 & \textbf{51.83} & \textbf{54.03} & \textbf{54.87}
& 58.58 & 45.46 & 43.30 & 43.07 \\
& EvoSkill
& 56.80 & 63.31 & 63.91 & 63.91
& 32.97 & 44.13 & 45.60 & 45.95
& 56.80 & 46.35 & 45.56 & 45.41 \\
& \cellcolor{oursblue}Ours
& \cellcolor{oursblue}\textbf{61.13}
& \cellcolor{oursblue}\textbf{72.19}
& \cellcolor{oursblue}\textbf{74.34}
& \cellcolor{oursblue}\textbf{75.47}
& \cellcolor{oursblue}\textbf{34.60}
& \cellcolor{oursblue}47.38
& \cellcolor{oursblue}50.44
& \cellcolor{oursblue}51.48
& \cellcolor{oursblue}\textbf{61.13}
& \cellcolor{oursblue}\textbf{50.44}
& \cellcolor{oursblue}\textbf{49.98}
& \cellcolor{oursblue}\textbf{49.73} \\

\addlinespace[2pt]
\midrule

\multirow{4}{*}{\textbf{Module}}
& Baseline
& 55.56 & 62.22 & 62.22 & 62.22
& 33.98 & 43.56 & 43.73 & 43.73
& 55.56 & 49.26 & 48.06 & 47.65 \\
& AutoSkill
& 46.99 & 55.42 & 56.63 & 56.63
& 36.12 & 45.65 & 45.96 & 45.96
& 46.99 & 43.37 & 42.87 & 42.71 \\
& EvoSkill
& 49.40 & 55.42 & 55.42 & 55.42
& 36.93 & 46.60 & 47.20 & 47.50
& 49.40 & 45.98 & 45.82 & 45.67 \\
& \cellcolor{oursblue}Ours
& \cellcolor{oursblue}\textbf{59.26}
& \cellcolor{oursblue}\textbf{66.67}
& \cellcolor{oursblue}\textbf{67.41}
& \cellcolor{oursblue}\textbf{67.41}
& \cellcolor{oursblue}\textbf{39.43}
& \cellcolor{oursblue}\textbf{46.90}
& \cellcolor{oursblue}\textbf{47.89}
& \cellcolor{oursblue}\textbf{48.29}
& \cellcolor{oursblue}\textbf{59.26}
& \cellcolor{oursblue}\textbf{53.33}
& \cellcolor{oursblue}\textbf{52.23}
& \cellcolor{oursblue}\textbf{51.97} \\

\addlinespace[2pt]
\midrule

\multirow{4}{*}{\textbf{Function}}
& Baseline
& 37.37 & 54.21 & 55.79 & 57.89
& 20.45 & 36.45 & 37.94 & 40.30
& 37.37 & 35.00 & 32.68 & 32.05 \\
& AutoSkill
& 39.29 & 51.79 & 55.36 & 56.25
& 24.19 & 32.35 & 34.34 & 32.97
& 39.29 & 28.87 & 27.99 & 27.67 \\
& EvoSkill
& 39.29 & 50.89 & 50.89 & 51.79
& 24.15 & 33.02 & 33.79 & 33.83
& 39.29 & 32.59 & 30.94 & 30.77 \\
& \cellcolor{oursblue}\textbf{Ours}
& \cellcolor{oursblue}\textbf{43.68}
& \cellcolor{oursblue}\textbf{58.95}
& \cellcolor{oursblue}\textbf{61.58}
& \cellcolor{oursblue}\textbf{62.63}
& \cellcolor{oursblue}\textbf{25.98}
& \cellcolor{oursblue}\textbf{38.63}
& \cellcolor{oursblue}\textbf{42.08}
& \cellcolor{oursblue}\textbf{43.05}
& \cellcolor{oursblue}\textbf{43.68}
& \cellcolor{oursblue}\textbf{36.84}
& \cellcolor{oursblue}\textbf{35.69}
& \cellcolor{oursblue}\textbf{34.92} \\

\bottomrule

\CodeAfter
\tikz \draw[line width=0.5pt] (1-|2) -- (last-|2);
\tikz \draw[line width=0.5pt] (1-|3) -- (last-|3);

\end{NiceTabular}
}
\vspace{-12pt}
\end{table*}

\subsection{Experimental Setup}
We evaluate SkillLens on MuLocbench~\citep{zhang2025MuLocbench} and ALFWorld~\citep{shridhar2021alfworld} using Codex GPT-5.2, and compare it with Vanilla, AutoSkill~\citep{yang2026autoskill}, and EvoSkill~\citep{alzubi2026evoskill} under matched agents, data splits, execution budgets, and task environments. Full model configurations, dataset construction, baseline definitions, and evolution and evaluation protocols are provided in Appendix~\ref{app:implementation-details}.

\begin{table}[htbp]
\centering
\small
\caption{ALFWorld results across different subtasks. We report the average success rate for each subtask as well as the overall result. Better values are in bold.}
\label{tab:valid_seen_task_success_rate}
\setlength{\tabcolsep}{5pt}
\renewcommand{\arraystretch}{1.05}

\resizebox{\columnwidth}{!}{
\begin{tabular}{cccccccc}
\toprule
\textbf{Method} & Heat & Pick2 & Clean & Cool & Look & Pick & Overall \\
\midrule
Baseline
& 43.75 & 50.00 & 25.93 & 16.00 & \textbf{76.92} & \textbf{65.71} & 45.00 \\
AutoSkill
& 37.50 & 50.00 & 37.04 & \textbf{24.00} & \textbf{76.92} & 62.86 & 47.14 \\
EvoSkill
& 31.25 & 50.00 & \textbf{40.74} & 20.00 & \textbf{76.92} & \textbf{65.71} & 47.14 \\
\rowcolor{oursblue}
\textbf{Ours}
& \textbf{50.00} & \textbf{54.17} & 37.04 & \textbf{24.00} & \textbf{76.92} & \textbf{65.71} & \textbf{51.31} \\
\bottomrule
\end{tabular}
}
\end{table}

\subsection{Main Results}

\noindent \textbf{SkillLens achieves superior performance across both benchmarks.}
On MuLocbench, SkillLens obtains the best Acc@1 at all three localization granularities, reaching 61.13\%, 59.26\%, and 43.68\% at the file, module, and function levels, respectively. On ALFWorld, it achieves a 51.31\% overall success rate, improving over Vanilla by 6.31 percentage points and over both AutoSkill and EvoSkill by 4.17 points.

\noindent \textbf{Advantages grow with task difficulty.}
On MuLocbench, the gain over Vanilla increases from 2.26 percentage points at the file level to 3.70 points at the module level and 6.31 points at the function level. On ALFWorld, simple subtasks such as Look and Pick are near saturation, whereas SkillLens reaches 50.00\% on Heat and 54.17\% on Pick2, outperforming all baselines. The pattern suggests that mixed-granularity adaptation is most useful for fine-grained or compositional tasks.

\noindent \textbf{Mixed granularity and dual-registry drive the gains.}
AutoSkill attains high file-level Recall@10 on MuLocbench (54.87\%) but lower Precision@10 (43.07\%), while EvoSkill reaches 47.14\% success on ALFWorld. SkillLens instead uses verifier-driven decomposition and local rewriting to preserve useful subunits, while dual-registry evolution jointly improves skill retrieval and routing.


\subsection{Ablation Studies}
\begin{table*}[t]
\centering
\small
\caption{MuLocBench ablation results for RWR across different granularity levels. All values are reported in percentage. Better values are in bold.}
\label{tab:rwr_ablation_results}
\vspace{3pt}
\setlength{\tabcolsep}{4.2pt}
\renewcommand{\arraystretch}{1.08}
\resizebox{\textwidth}{!}{
\begin{NiceTabular}{cccccc|cccc|cccc}
\toprule
\multirow{2}{*}{\textbf{Granularity}}
& \multirow{2}{*}{\textbf{Method}}
& \multicolumn{4}{c|}{\textbf{Accuracy}}
& \multicolumn{4}{c|}{\textbf{Recall}}
& \multicolumn{4}{c}{\textbf{Precision}} \\
\cmidrule(r){3-6} \cmidrule(lr){7-10} \cmidrule(l){11-14}
& & @1 & @3 & @5 & @10
& @1 & @3 & @5 & @10
& @1 & @3 & @5 & @10 \\
\midrule

\multirow{2}{*}{\textbf{File}}
& w/o RWR
& 60.00 & 68.68 & 71.32 & 73.21
& 32.53 & 43.68 & 46.50 & 48.22
& 60.00 & 48.87 & 48.95 & 48.77 \\
& \cellcolor{oursblue}Ours
& \cellcolor{oursblue}\textbf{61.13}
& \cellcolor{oursblue}\textbf{72.19}
& \cellcolor{oursblue}\textbf{74.34}
& \cellcolor{oursblue}\textbf{75.47}
& \cellcolor{oursblue}\textbf{34.60}
& \cellcolor{oursblue}\textbf{47.38}
& \cellcolor{oursblue}\textbf{50.44}
& \cellcolor{oursblue}\textbf{51.48}
& \cellcolor{oursblue}\textbf{61.13}
& \cellcolor{oursblue}\textbf{50.44}
& \cellcolor{oursblue}\textbf{49.98}
& \cellcolor{oursblue}\textbf{49.73} \\

\addlinespace[2pt]
\midrule

\multirow{2}{*}{\textbf{Module}}
& w/o RWR
& 54.07 & 63.71 & 65.93 & 65.19
& 34.05 & 45.01 & 47.56 & 47.86
& 54.07 & 49.87 & 49.19 & 48.80 \\
& \cellcolor{oursblue}Ours
& \cellcolor{oursblue}\textbf{59.26}
& \cellcolor{oursblue}\textbf{66.67}
& \cellcolor{oursblue}\textbf{67.41}
& \cellcolor{oursblue}\textbf{67.41}
& \cellcolor{oursblue}\textbf{39.43}
& \cellcolor{oursblue}\textbf{46.90}
& \cellcolor{oursblue}\textbf{47.89}
& \cellcolor{oursblue}\textbf{48.29}
& \cellcolor{oursblue}\textbf{59.26}
& \cellcolor{oursblue}\textbf{53.33}
& \cellcolor{oursblue}\textbf{52.23}
& \cellcolor{oursblue}\textbf{51.97} \\

\addlinespace[2pt]
\midrule

\multirow{2}{*}{\textbf{Function}}
& w/o RWR
& 36.31 & 54.74 & 58.42 & 60.00
& 20.64 & 34.64 & 38.88 & 40.03
& 36.31 & 34.03 & 33.11 & 32.22 \\
& \cellcolor{oursblue}\textbf{Ours}
& \cellcolor{oursblue}\textbf{43.68}
& \cellcolor{oursblue}\textbf{58.95}
& \cellcolor{oursblue}\textbf{61.58}
& \cellcolor{oursblue}\textbf{62.63}
& \cellcolor{oursblue}\textbf{25.98}
& \cellcolor{oursblue}\textbf{38.63}
& \cellcolor{oursblue}\textbf{42.08}
& \cellcolor{oursblue}\textbf{43.05}
& \cellcolor{oursblue}\textbf{43.68}
& \cellcolor{oursblue}\textbf{36.84}
& \cellcolor{oursblue}\textbf{35.69}
& \cellcolor{oursblue}\textbf{34.92} \\

\bottomrule

\CodeAfter
\tikz \draw[line width=0.5pt] (1-|2) -- (last-|2);
\tikz \draw[line width=0.5pt] (1-|3) -- (last-|3);

\end{NiceTabular}
}
\end{table*}

\begin{table*}[t]
\centering
\small
\caption{MuLocbench ablation results under different partial-rewrite strategies. All values are reported in percentage. Better values are in bold.}
\label{tab:partial_rewrite_full_metrics_aligned}
\setlength{\tabcolsep}{4.2pt}
\renewcommand{\arraystretch}{1.08}
\resizebox{\textwidth}{!}{
\begin{NiceTabular}{cccccc|cccc|cccc}
\toprule
\multirow{2}{*}{\textbf{Granularity}}
& \multirow{2}{*}{\textbf{Method}}
& \multicolumn{4}{c|}{\textbf{Accuracy}}
& \multicolumn{4}{c|}{\textbf{Recall}}
& \multicolumn{4}{c}{\textbf{Precision}} \\
\cmidrule(r){3-6} \cmidrule(lr){7-10} \cmidrule(l){11-14}
& & @1 & @3 & @5 & @10
& @1 & @3 & @5 & @10
& @1 & @3 & @5 & @10 \\
\midrule

\multirow{3}{*}{\textbf{File}}
& Parent-only
& \textbf{61.51} & 70.68 & 73.58 & 73.96
& \textbf{35.45} & 46.64 & \textbf{50.45} & 50.99
& \textbf{61.51} & 50.00 & 49.77 & 49.49 \\
& Rewrite-all
& 60.37 & 70.68 & \textbf{74.34} & 74.71
& 33.97 & 45.93 & 49.98 & 50.78
& 60.37 & 48.87 & 49.15 & 48.71 \\
& \cellcolor{oursblue}\textbf{Ours}
& \cellcolor{oursblue}61.13
& \cellcolor{oursblue}\textbf{72.19}
& \cellcolor{oursblue}\textbf{74.34}
& \cellcolor{oursblue}\textbf{75.47}
& \cellcolor{oursblue}34.60
& \cellcolor{oursblue}\textbf{47.38}
& \cellcolor{oursblue}50.44
& \cellcolor{oursblue}\textbf{51.48}
& \cellcolor{oursblue}61.13
& \cellcolor{oursblue}\textbf{50.44}
& \cellcolor{oursblue}\textbf{49.98}
& \cellcolor{oursblue}\textbf{49.73} \\

\addlinespace[2pt]
\midrule

\multirow{3}{*}{\textbf{Module}}
& Parent-only
& 57.78 & 65.93 & 66.67 & 66.67
& \textbf{39.43} & \textbf{47.05} & 47.42 & 47.96
& 57.78 & 51.55 & 50.05 & 49.96 \\
& Rewrite-all
& 55.56 & 64.45 & 65.93 & 65.93
& 36.28 & 44.99 & 47.25 & 47.30
& 55.56 & 51.10 & 49.70 & 49.50 \\
& \cellcolor{oursblue}\textbf{Ours}
& \cellcolor{oursblue}\textbf{59.26}
& \cellcolor{oursblue}\textbf{66.67}
& \cellcolor{oursblue}\textbf{67.41}
& \cellcolor{oursblue}\textbf{67.41}
& \cellcolor{oursblue}\textbf{39.43}
& \cellcolor{oursblue}46.90
& \cellcolor{oursblue}\textbf{47.89}
& \cellcolor{oursblue}\textbf{48.29}
& \cellcolor{oursblue}\textbf{59.26}
& \cellcolor{oursblue}\textbf{53.33}
& \cellcolor{oursblue}\textbf{52.23}
& \cellcolor{oursblue}\textbf{51.97} \\

\addlinespace[2pt]
\midrule

\multirow{3}{*}{\textbf{Function}}
& Parent-only
& \textbf{45.79} & 58.42 & 61.05 & 61.05
& 25.98 & 38.55 & 41.92 & 42.86
& \textbf{45.79} & 36.70 & 35.48 & 34.70 \\
& Rewrite-all
& 45.26 & 56.84 & 60.00 & 60.53
& 25.70 & 37.43 & 40.66 & 42.23
& 45.26 & 35.53 & 33.84 & 32.88 \\
& \cellcolor{oursblue}\textbf{Ours}
& \cellcolor{oursblue}43.68
& \cellcolor{oursblue}\textbf{58.95}
& \cellcolor{oursblue}\textbf{61.58}
& \cellcolor{oursblue}\textbf{62.63}
& \cellcolor{oursblue}\textbf{26.12}
& \cellcolor{oursblue}\textbf{38.63}
& \cellcolor{oursblue}\textbf{42.08}
& \cellcolor{oursblue}\textbf{43.05}
& \cellcolor{oursblue}43.68
& \cellcolor{oursblue}\textbf{36.84}
& \cellcolor{oursblue}\textbf{35.69}
& \cellcolor{oursblue}\textbf{34.92} \\

\bottomrule

\CodeAfter
\tikz \draw[line width=0.5pt] (1-|2) -- (last-|2);
\tikz \draw[line width=0.5pt] (1-|3) -- (last-|3);

\end{NiceTabular}
}
\vspace{-15pt}
\end{table*}
\noindent \textbf{Effect of Degree-Corrected Random Walk.} To isolate the contribution of degree-corrected RWR, we compare the full pipeline against a variant that uses only embedding-based top-$K$ retrieval as the skill source (w/o RWR), while keeping all subsequent components including verifier-driven traversal identical. This ablation directly measures whether propagating seed signals through the skill graph improves retrieval quality beyond flat embedding similarity.

\looseness=-1
Table~\ref{tab:rwr_ablation_results} shows that RWR consistently improves performance across all three granularities, with gains that correlate with graph structural density. At the module level, which serves as the structural bottleneck of the four-layer hierarchy with the densest hierarchical and lateral edges, RWR delivers the largest relative improvement: Acc@1 rises from 54.07\% to 59.26\% (+5.19 pp) while Acc@5 increases only modestly (+1.48 pp). This pattern indicates that RWR primarily re-ranks candidates rather than merely adding new ones—the correct module already exists in the top-5 but is promoted to the top position through graph propagation. At the function level, the most fine-grained and challenging setting, RWR achieves the largest absolute gain (Acc@1 +7.37 pp, Recall@1 +5.34 pp), demonstrating its ability to surface procedurally relevant units that are textually distant from the query but structurally reachable via hierarchical edges. File-level gains are smallest, consistent with the sparser connectivity among file nodes in the skill graph. Across all granularities, precision does not degrade despite increased recall, confirming that degree correction successfully prevents the walk from collapsing into dense similarity clusters that would otherwise introduce noise. These results validate that propagating relevance along the skill graph yields better-ranked and more complete skill contexts than flat embedding retrieval alone.

\begin{figure}[t]
    \centering
    \includegraphics[width=\linewidth]{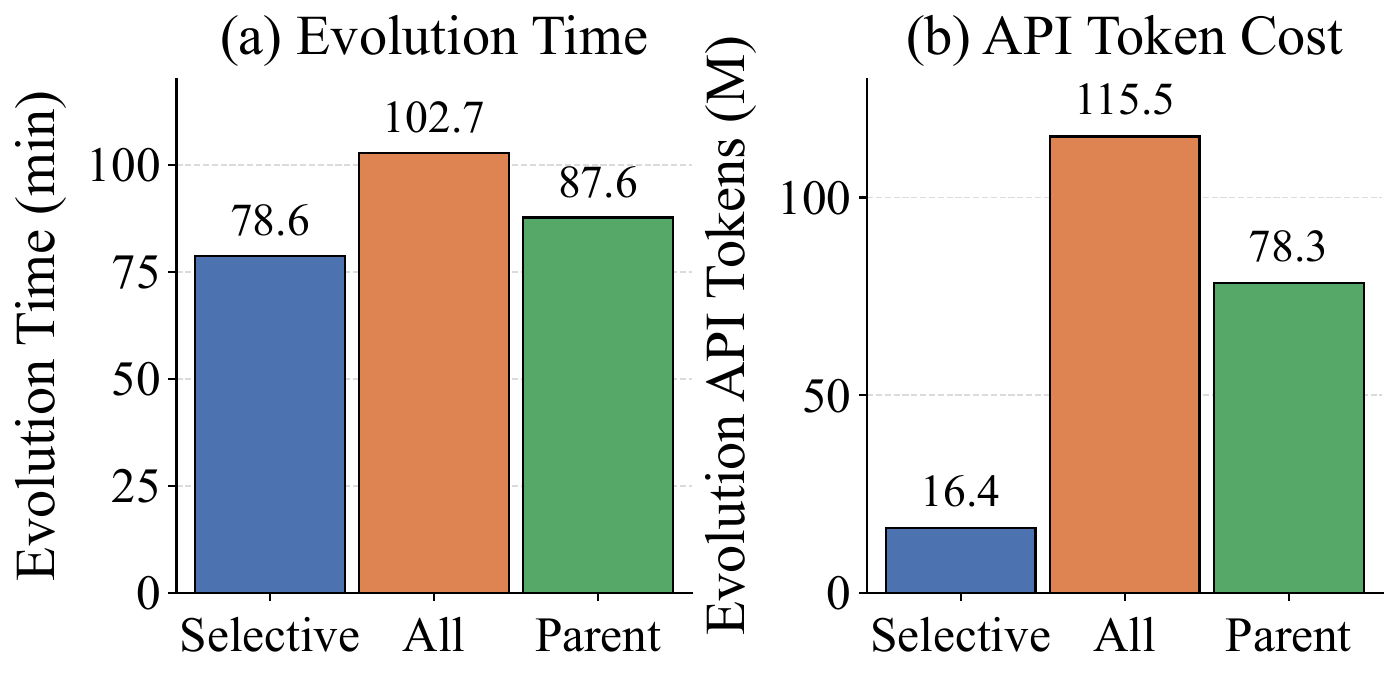}
    \caption{Evolution computation cost of three re-writing methods.}
    \label{fig:rewrite_compare}
\end{figure}

\noindent \textbf{Effect of different partial-rewrite strategies.} To evaluate the effectiveness and efficiency of our selective rewriting strategy, where we perform subtree drill-down on partial skill nodes and selectively rewrite
only a subset of their child skill nodes, we compare it with two baselines:

(i) rewrite all child skill nodes, which skips the verification step and rewrites all child skill nodes before merging them; and (ii) rewrite parent skill nodes,
which rewrites only the partial skill nodes that serve as parent nodes.

Selective drill-down rewriting improves localization coverage while reducing evolution cost. 
Compared with Parent-only and Rewrite-all from Table~\ref{tab:partial_rewrite_full_metrics_aligned}, our method mainly improves the completeness of the returned candidate set. 
At the file level, although Parent-only slightly leads on Acc@1 (61.51\% vs. 61.13\%), our method achieves higher Acc@3/10 (72.19\%/75.47\%) than Parent-only (70.68\%/73.96\%) and Rewrite-all (70.68\%/74.71\%). 
At the module level, our method obtains the best Acc@1--10, exceeding Parent-only and Rewrite-all by 1.48 and 3.70 points on Acc@1. 
At the function level, Parent-only has higher Acc@1 (45.79\% vs. 43.68\%), but our method achieves the best Acc@3/5/10 (58.95\%/61.58\%/62.63\%) and Recall@1/3/5/10 (26.12\%/38.63\%/42.08\%/43.05\%). 
These gains are not due to a larger rewriting budget: Rewrite-all costs 6,161.83 seconds and 115.51M tokens, Parent-only costs 5,257.82 seconds and 78.35M tokens, while our method uses only 4,718.54 seconds and 16.41M tokens. 
This shows that drilling down before rewriting preserves compatible subskills, avoids unnecessary edits, and yields a more task-specific skill context.

\subsection{Computational Cost Analysis}

\begin{figure}[t]
    \centering
    \includegraphics[width=\columnwidth]{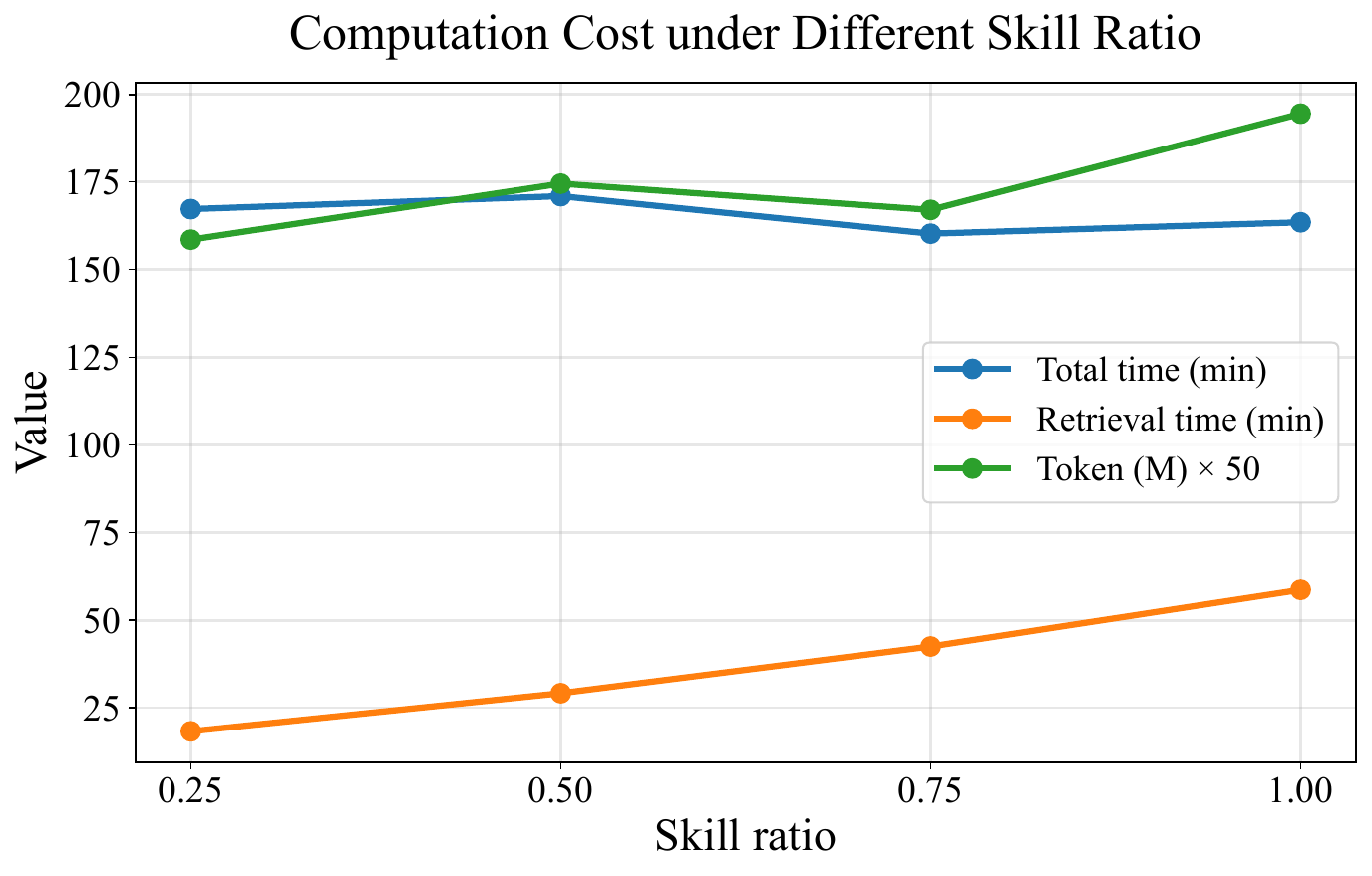}
    \caption{Computation cost under different initial skill ratios.}
    \label{fig:computation_cost}
\end{figure}

To analyze the impact of the number of initialized skills on computational cost, we conduct comparative experiments under different initial skill ratios, ranging from 25\% to 100\%. Specifically, we measure the token consumption of the main agent model, as well as the time spent on skill retrieval and skill evolution.

\looseness=-1
As shown in Figure~\ref{fig:computation_cost}, increasing the initial skill ratio mainly increases retrieval overhead. 
When the skill ratio grows from 25\% to 100\%, total retrieval time rises from 18.28 to 58.72 minutes, a 3.21$\times$ increase. 
Main-agent token usage also increases, but more moderately, from 3.17M to 3.89M tokens, corresponding to a 22.7\% increase. 
In contrast, total evolution time is not monotonic across skill ratios, suggesting that wall-clock cost is also affected by parallel execution, the number of triggered refinement steps, and per-instance runtime variation. 
Overall, scaling the initialized skill pool primarily increases retrieval cost, while token growth and total evolution time remain comparatively less sensitive.

\section{Conclusion}

\looseness=-1
This work examines a key limitation of existing skill-based LLM agents: retrieved skills are usually reused as flat prompt blocks. 
When a skill is only partially relevant, this all-or-nothing design can introduce noise, waste reusable substructure, or require unnecessary rewriting. Motivated by this problem, we introduce SkillLens, a framework that organizes skills hierarchically and constructs task-specific skill contexts at the appropriate resolution. Experiments on MuLocbench and ALFWorld show that SkillLens outperforms strong evolving-skill baselines, with larger gains on fine-grained and compositional tasks. Future work will extend SkillLens to multimodal tool-use, web navigation, and multi-agent settings, while improving skill compression for large evolving libraries.

\section*{Limitations}

Our evaluation covers two task families: repository-level bug localization and
interaction in a text-based embodied environment. Both benchmarks are
English-centric and use fixed task protocols. The results therefore support the
effectiveness of mixed-granularity skill reuse in these settings, but do not by
themselves establish the same benefits for other languages, multimodal agents,
or safety-critical deployments. Likewise, benchmark task distributions may not
capture the full diversity of open-ended agent use.

SkillLens also depends on the quality of its initial skill library and on the
verifier's routing decisions. Noisy or poorly structured skills can still lead
to unnecessary decomposition or rewriting, and the four-level ontology used in
this work may not be the most suitable organization for every domain. In
addition, our experiments use one agent--writer model family, one verifier
model, and one embedding model; the magnitude of the gains may vary with other
model and retrieval choices.

Finally, hierarchical retrieval and registry evolution introduce verifier calls
and maintenance overhead relative to an agent without external skills. Although
we analyze computation under different initialization ratios, we do not study
continual library growth over very long deployments. Our theoretical results
also rely on sparse-mismatch and retrieval-calibration assumptions. They
characterize the intended operating regime of SkillLens rather than guarantee
identical improvements for arbitrary skill graphs or task distributions.

\bibliography{custom}

\appendix

\section{Implementation Details}
\label{app:implementation-details}

\subsection{Models and Initialization}
We use Codex GPT-5.2 for the agent \(\mathcal{A}_{\theta}\) and skill writer \(\mathcal{W}_{\phi}\), GPT-5-mini for the verifier, and text-embedding-3-small for retrieval and graph initialization. For both benchmarks, the evolution and validation sets are split 7:3, with held-out test sets. The skill library is initialized with antigravity-awesome-skills.

\subsection{Datasets and Evaluation}
For repository-localization tasks, the training dataset contains 36 repositories: AutoGPT, ChatGLM-6B, ComfyUI, Deep-Live-Cam, DeepSeek-V3, Fooocus, MetaGPT, PaddleOCR, Real-Time-Voice-Cloning, ansible, core, cpython, face\_recognition, faceswap, fastapi, flask, gpt-engineer, gpt4free, gpt\_academic, grok-1, hackingtool, langflow, llama, manim, odoo, open-interpreter, private-gpt, pytorch, rich, scikit-learn, scrapy, thefuck, transformers, yolov5, youtube-dl, and yt-dlp. The test dataset contains 10 repositories: LLaMA-Factory, OpenHands, keras, localstack, pandas, requests, screenshot-to-code, sherlock, stable-diffusion-webui, and text-generation-webui. We evaluate on MuLocbench, where the agent predicts ranked file-, module-, and function-level candidates for each issue, and report accuracy, recall, and precision at all three granularities.

For text-environment tasks, we evaluate ALFWorld with full-episode rollouts of at most 50 steps. We report episode success rate (\texttt{won}) as the primary metric together with per-subtask success rates. The training split is used for evolution and validation, and the valid-seen split is reserved for testing.

\subsection{Baselines}
We compare SkillLens with three baselines: (i) \textbf{Vanilla}, the base Codex GPT-5.2 agent without external skills; (ii) \textbf{AutoSkill}~\citep{yang2026autoskill}, a flat skill self-evolution method that abstracts reusable skills from interaction traces; and (iii) \textbf{EvoSkill}~\citep{alzubi2026evoskill}, a failure-driven method that refines skills from failed executions. All methods use the same underlying agent, data splits, execution budget, and task environments; they differ only in how skills are constructed, retrieved, adapted, and evolved.

\subsection{Evolution and Evaluation Protocol}
For both repository-localization and text-environment experiments, we first run the base agent \(\mathcal{A}_{\theta}\) on the evolution split and construct gap reports from failed cases. Each report records the error type, failed prediction or trajectory, ground-truth target, retrieved skills, verifier routing decisions, and invoked or rewritten skill units. The reports are then used to update the skill graph and refine retrievable skill units. After each iteration, we evaluate the updated system on the validation split and select the best version for a single final test-set evaluation.

All evaluations are executed in reproducible isolated environments. For code-agent tasks, each issue is run in a repository-specific sandbox at the target code snapshot. For ALFWorld, each episode uses an independent TextWorld/ALFWorld instance with a fixed step budget.


\section{Proofs}
\label{app:proofs}

\subsection{Proof of Proposition 1 (Sublinear cost of adaptation)}
\label{app:proof-prop1}

Each \textsc{Accept} or \textsc{Skip} decision terminates exploration of the corresponding subtree, 
so the visited set is exactly the adaptive frontier together with its ancestors. 
The expected size of the visited frontier at depth \(d\) satisfies \(\mathbb{E}[|F_d|]\le (\rho b)^{d}\), and summing along the tree yields
\begin{equation}
\begin{aligned}
\mathbb{E}[N_{\mathrm{vis}}]&\le\sum_{d=0}^{D}(\rho b)^{d}=O\!\Big(\frac{1}{1-\rho b}\Big)=O(1),\\
\mathbb{E}[N_{\mathrm{rw}}]&\le \mathbb{E}[N_{\mathrm{vis}}].
\end{aligned}
\end{equation}
Hence \(\mathrm{C}_{\mathrm{adapt}}=O(N_{\mathrm{vis}}+N_{\mathrm{rw}})\) is independent of \(n\) in expectation, and in the worst case (\(\rho b\to 1\)) is bounded by \(O(D)=O(\log n)\). In both regimes \(\mathrm{C}_{\mathrm{adapt}}=o(n)\), in contrast to flat full-skill injection whose token cost is \(\Theta(n)\). \(\square\)

\subsection{Proof of Proposition 2 (Monotone improvement and convergence to a local optimum)}
\label{app:proof-prop2}

Monotonicity follows directly from including the previous pair in the candidate set, 
which guarantees \(J^{(t+1)}\ge J^{(t)}\). Since \(M\) is bounded (as a task metric in \([0,1]\)) 
and \(\mathrm{C}\) is bounded below, \(J\) is bounded above, so \(\{J^{(t)}\}\) is monotone and bounded 
and hence convergent. At the limit, by definition of the update rule no atomic edit in \(\mathcal{P}_{\infty}\) 
yields strict improvement, so \(\mathcal{S}^{\dagger}\) 
is a fixed point of the operator family, i.e.\ a local optimum under the neighborhood induced by single atomic edits. 
The gap \(J^{*}-J^{\dagger}\) is in general non-zero because the search proceeds via finite atomic neighborhoods rather 
than a global combinatorial sweep over all skill-pair configurations; this yields the claimed sub-optimality with respect to the evolution objective. \(\square\)

\subsection{Proof of Proposition 3 (Composed approximation guarantee)}
\label{app:proof-prop3}

The three stages of Algorithm~\ref{alg:adaptation} approximate the surrogate \(J_Q\) block-by-block, and their errors compose additively under assumptions (a)--(d).

\emph{Step 1 (retrieval block).}
Top-\(K\) embedding retrieval reduces to a singleton-relevance subproblem
\(\max_{R:|R|\le K}\sum_{u\in R}r_Q(u)\). Since \(r_Q\) calibrates \(F_Q(\{u\})\) up to \(\epsilon_{\mathrm{ret}}\), the returned seed set \(\widehat{R}_0\) satisfies
\(\sum_{u\in\widehat{R}_0}F_Q(\{u\})\ge\max_{R}\sum_{u\in R}F_Q(\{u\})-\epsilon_{\mathrm{ret}}\), i.e.\ optimal up to \(\epsilon_{\mathrm{ret}}\).

\emph{Step 2 (random-walk expansion).}
Under assumption (b), \(F_Q\) is monotone submodular, so the classical greedy bound applies to any selection rule that picks elements in order of estimated marginal gain. By assumption (c), the degree-corrected RWR score \(s(u)\) approximates the true marginal gain \(\Delta(u\mid R)\) up to \(\epsilon_{\mathrm{rwr}}\); standard noisy-greedy analysis~\citep{nemhauser1978submodular} then yields
\(J_Q^{\mathrm{ret}}(\widehat{R})\ge(1-1/e)\,J_Q^{\mathrm{ret}}(R^{*})-\epsilon_{\mathrm{rwr}}\),
inheriting the \((1-1/e)\) submodular guarantee on the retrieval block.

\emph{Step 3 (verifier-driven rewrite).}
Because the writer \(\mathcal{W}_{\phi}\) operates on each unit independently, the rewrite block is additive over routed nodes: \(J_Q^{\mathrm{rw}}(B)=\sum_{u\in\widehat{R}}\phi(a_u)\) for per-node action utilities \(\phi(\cdot)\). Under assumption (d), the verifier matches the optimal local action of Eq.~\eqref{eq:routing} up to per-node error \(\epsilon_{\mathrm{ver}}\), so summing over the \(|\widehat{R}|\) routed nodes gives
\(J_Q^{\mathrm{rw}}(\widehat{B})\ge J_Q^{\mathrm{rw}}(B^{*})-|\widehat{R}|\,\epsilon_{\mathrm{ver}}\).
Step 3 is exact when \(\epsilon_{\mathrm{ver}}=0\) and otherwise degrades linearly in the number of routed nodes.

\emph{Composition.}
Since the surrogate decomposes additively as \(J_Q=J_Q^{\mathrm{ret}}+J_Q^{\mathrm{rw}}\), summing the three per-block bounds yields Eq.~\eqref{eq:composed-bound}. The procedure is therefore not an exact optimizer of the unobservable correctness metric \(M\), but an objective-consistent approximation to its tractable surrogate, with each step playing a clean role in Eq.~\eqref{eq:objective}. \(\square\)

\section{Case Study}
\label{app:case-study}

\subsection{MuLocbench Example: YOLOv5 Root-Path Failure}

\paragraph{Issue.}
Repository: \texttt{ultralytics/yolov5}.  
The user reports that loading a custom model through Torch Hub fails:

\begin{tcolorbox}[
  colback=caseblue,
  colframe=blue!35!black,
  title=\textbf{Problem instance},
  fonttitle=\bfseries,
  arc=1.5mm,
  boxrule=0.6pt,
  breakable
]
\small
\textbf{Command:}
\begin{quote}
\ttfamily\raggedright
torch.hub.load(\allowbreak 'ultralytics/yolov5', 'custom', path='yolov5/path/last.pt', force\_reload=True)
\end{quote}

\textbf{Observed error:}
\begin{quote}
\ttfamily\raggedright
ValueError: `C:\textbackslash\allowbreak Users\textbackslash\allowbreak aaa\textbackslash\allowbreak .cache\textbackslash\allowbreak torch\textbackslash\allowbreak hub\textbackslash\allowbreak ultralytics\_yolov5\_master' \\
does not start with \\
`C:\textbackslash\allowbreak Users\textbackslash\allowbreak aaa\textbackslash\allowbreak PycharmProjects\textbackslash\allowbreak project\textbackslash\allowbreak proejct1'
\end{quote}

This indicates a mismatch between the repository root resolved inside the cached Torch Hub checkout and the user's current working directory.
\end{tcolorbox}

\begin{tcolorbox}[
  colback=casegray,
  colframe=black!25,
  title=\textbf{Retrieved skill},
  fonttitle=\bfseries,
  arc=1.5mm,
  boxrule=0.6pt,
  breakable
]
\small
\textbf{Skill: External-loader path-root localization}

\medskip
\textbf{Policy.}
If a Python project fails only when loaded through an external mechanism such as
\texttt{torch.hub}, a package cache, or a plugin loader, first inspect how the code computes the repository root and whether that root is forcibly converted relative to \texttt{Path.cwd()}.

\medskip
\textbf{Procedure.}
\begin{itemize}
    \item Trace the reported execution entry and identify the first repository modules imported on that path.
    \item Search these modules for root-path construction with \texttt{Path(\_\_file\_\_)}, \texttt{Path.cwd()}, and \texttt{relative\_to}.
    \item If the error says one path ``does not start with'' another, prioritize direct calls to \texttt{ROOT.relative\_to(Path.cwd())}.
    \item Prefer a guarded fallback: keep relative paths when valid, but preserve the absolute root when the repository is outside the current working directory.
\end{itemize}
\end{tcolorbox}

\begin{tcolorbox}[
  colback=casegreen,
  colframe=green!40!black,
  title=\textbf{Drill-down result},
  fonttitle=\bfseries,
  arc=1.5mm,
  boxrule=0.6pt,
  breakable
]
\small
\textbf{Kept context.}
\begin{itemize}
    \item The reported command uses \texttt{torch.hub.load(..., 'custom', ...)}.
    \item Therefore, the primary search path is the Hub loading path rather than standalone training or inference scripts.
    \item The error pattern strongly suggests failure in repository-root normalization.
    \item The most likely problematic operation is \texttt{ROOT.relative\_to(Path.cwd())}.
\end{itemize}

\textbf{Task-specific focus.}
\begin{itemize}
    \item Inspect the modules directly involved in Hub-based model loading.
    \item Prioritize \texttt{models/yolo.py}.
    \item Treat \texttt{models/tf.py}, \texttt{detect.py}, \texttt{train.py}, \texttt{val.py}, and \texttt{export.py} as secondary locations only if they reuse the same fragile pattern.
\end{itemize}
\end{tcolorbox}

\begin{tcolorbox}[
  colback=white,
  colframe=black!35,
  title=\textbf{Final localization},
  fonttitle=\bfseries,
  arc=1.5mm,
  boxrule=0.6pt,
  breakable
]
\small
With the refined context, the agent localizes the issue to YOLOv5's root-path conversion logic.

\medskip
\textbf{Primary file:}
\begin{itemize}
    \item \texttt{models/yolo.py}
\end{itemize}

\textbf{Secondary consistency files:}
\begin{itemize}
    \item \texttt{models/tf.py}
    \item \texttt{detect.py}
    \item \texttt{train.py}
    \item \texttt{val.py}
    \item \texttt{export.py}
\end{itemize}

\textbf{Localized fix.}
Guard the root conversion
\texttt{ROOT = ROOT.relative\_to(Path.cwd())}
with \texttt{try/except ValueError}. If the repository is loaded from the Torch Hub cache and is not under the user's current working directory, keep \texttt{ROOT} as an absolute path instead of raising an exception.
\end{tcolorbox}

\begin{tcolorbox}[
  colback=caseorange,
  colframe=orange!60!black,
  title=\textbf{Gap Analysis},
  fonttitle=\bfseries,
  arc=1.5mm,
  boxrule=0.6pt,
  breakable
]
\small
The retrieved skill correctly identifies the failure pattern as an unsafe conversion from an absolute repository root to a path relative to \texttt{Path.cwd()} under an external loader.

However, the initial localization remains broader than necessary: it expands from the Hub loading path to several additional files that share similar root-handling logic. For this issue, the reported command primarily exercises the Torch Hub model-loading path, so \texttt{models/yolo.py} should receive the highest priority.
\end{tcolorbox}

\begin{tcolorbox}[
  colback=casegray,
  colframe=black!20,
  title=\textbf{Updated skill fragment},
  fonttitle=\bfseries,
  arc=1mm,
  boxrule=0.5pt,
  breakable
]
\small
When a repository behaves differently under an external loader than in direct execution, treat path-root resolution as a first-class localization target. Inspect how the project derives its repository root, how that root is normalized, and whether absolute paths are implicitly assumed to be inside the caller's current working directory.

Prioritize files that lie on the failing loader's actual import or execution path before expanding to other modules with similar path-handling code. Prefer guarded root-normalization logic: use relative paths only when the path relationship is valid, and otherwise preserve the absolute root so externally cached or relocated repositories remain loadable.
\end{tcolorbox}

\subsection{ALFWorld Example: Mixed-Granularity Skill Reuse}
\label{app:case-study-alfworld}

\begin{tcolorbox}[
  colback=caseblue,
  colframe=blue!35!black,
  title=\textbf{Problem instance},
  fonttitle=\bfseries,
  arc=1.5mm,
  boxrule=0.6pt,
  breakable
]
\small
\textbf{Goal:} Put a cooled apple on the dining table.

\medskip
\textbf{Environment context:}
The agent starts in a kitchen-like environment. The apple may be inside a drawer, cabinet, or on a countertop. 
To complete the task, the agent must first obtain the apple, cool it using a refrigerator, and then place it on the dining table.
\end{tcolorbox}

\begin{tcolorbox}[
  colback=casegray,
  colframe=black!25,
  title=\textbf{Retrieved skill: Clean object then place},
  fonttitle=\bfseries,
  arc=1.5mm,
  boxrule=0.6pt,
  breakable
]
\small
\textbf{Policy.}
For object-state-and-placement tasks, separate the plan into three reusable stages:
find and pick up the target object, transform the object's state using the correct appliance, and place the object at the requested destination.

\medskip
\textbf{Strategy.}
\begin{itemize}[leftmargin=12pt, itemsep=1pt, topsep=1pt]
    \item Search likely receptacles until the target object is found.
    \item Pick up the object before applying any state-changing operation.
    \item Use the environment object that matches the requested state change.
    \item After the state change is complete, navigate to the target receptacle and place the object there.
\end{itemize}

\medskip
\textbf{Stored procedure from prior task.}
\begin{itemize}[leftmargin=12pt, itemsep=1pt, topsep=1pt]
    \item Locate the apple.
    \item Pick up the apple.
    \item Clean the apple in the sink.
    \item Place the apple on the dining table.
\end{itemize}
\end{tcolorbox}

\begin{tcolorbox}[
  colback=casegreen,
  colframe=green!40!black,
  title=\textbf{Drill-down result},
  fonttitle=\bfseries,
  arc=1.5mm,
  boxrule=0.6pt,
  breakable
]
\small
\textbf{Kept units.}
\begin{itemize}[leftmargin=12pt, itemsep=1pt, topsep=1pt]
    \item Search likely receptacles for the apple.
    \item Pick up the apple after it is found.
    \item Navigate to the dining table after the state change.
    \item Put the apple on the dining table.
\end{itemize}

\textbf{Rewritten unit.}
\begin{itemize}[leftmargin=12pt, itemsep=1pt, topsep=1pt]
    \item Replace ``clean the apple in the sink'' with ``cool the apple in the fridge''.
\end{itemize}

\textbf{Dropped units.}
\begin{itemize}[leftmargin=12pt, itemsep=1pt, topsep=1pt]
    \item Sink-specific actions such as opening the sink, placing the apple in the sink basin, or using water are not included.
\end{itemize}
\end{tcolorbox}

\begin{tcolorbox}[
  colback=casegreen,
  colframe=green!40!black,
  title=\textbf{Constructed task-specific context},
  fonttitle=\bfseries,
  arc=1.5mm,
  boxrule=0.6pt,
  breakable
]
\small
For a ``cooled object then place'' task, first locate and pick up the target object. 
Then navigate to the refrigerator, open it, put the object inside, close it if needed, reopen it, and take the cooled object back. 
Finally, navigate to the requested destination and place the object there. 
Do not use sink-specific cleaning actions unless the task explicitly asks for a cleaned object.
\end{tcolorbox}

\begin{tcolorbox}[
  colback=white,
  colframe=black!35,
  title=\textbf{Final execution plan},
  fonttitle=\bfseries,
  arc=1.5mm,
  boxrule=0.6pt,
  breakable
]
\small
\begin{enumerate}[leftmargin=14pt, itemsep=1pt, topsep=1pt]
    \item Search likely locations such as the countertop, cabinet, and drawer.
    \item Take the apple once it is found.
    \item Go to the refrigerator.
    \item Open the refrigerator.
    \item Put the apple in the refrigerator.
    \item Close the refrigerator if the environment requires it.
    \item Open the refrigerator again and take the cooled apple.
    \item Go to the dining table.
    \item Put the cooled apple on the dining table.
\end{enumerate}
\end{tcolorbox}

\begin{tcolorbox}[
  colback=caseorange,
  colframe=orange!60!black,
  title=\textbf{Gap Analysis},
  fonttitle=\bfseries,
  arc=1.5mm,
  boxrule=0.6pt,
  breakable
]
\small
\textbf{Successful transfer.}
The retrieved skill correctly provides the high-level structure for object-state-and-placement tasks:
find the object, transform its state, and place it at the requested destination.

\textbf{Local mismatch.}
The retrieved procedure contains a cleaning operation, but the current task requires cooling. 
The mismatch is isolated to the state-transformation step rather than the whole skill.

\textbf{Agent-side update.}
Store a reusable procedure for cooled-object placement:
use the refrigerator as the default appliance for cooling tasks, then return the object to the requested destination.

\textbf{Verifier-side update.}
When a retrieved skill matches the object and destination but has a different state transformation, keep the acquisition and placement units, but rewrite only the state-changing procedure.
\end{tcolorbox}

\begin{tcolorbox}[
  colback=casegray,
  colframe=black!25,
  title=\textbf{Updated skill fragment},
  fonttitle=\bfseries,
  arc=1.5mm,
  boxrule=0.6pt,
  breakable
]
\small
For ALFWorld object-state-and-placement tasks, compare the requested state transformation before reusing a prior skill. 
If the object and destination match but the transformation differs, preserve the locate, pick, navigate, and place procedures, and rewrite only the transformation step. 
Use the sink for cleaning, the refrigerator for cooling, and the microwave or stove for heating, depending on the task instruction.
\end{tcolorbox}
\end{document}